\documentclass[runningheads]{llncs}
\usepackage[T1]{fontenc}
\usepackage{marvosym}
\usepackage{xcolor}
\usepackage{graphicx}
\usepackage{multirow}
\begin{document}
\title{Go-Explore for Residential Energy Management\thanks{Supported by Irish Research Council \& University of Galway}}
\author{Junlin Lu\inst{1}
% \orcidID{0000-0002-6014-9419} 
\and
Patrick Mannion\inst{1}
% \orcidID{0000-0002-7951-878X} 
\and
Karl Mason\inst{1}
% \orcidID{0000-0002-8966-9100}
}
\authorrunning{J. Lu, P. Mannion, K. Mason}
\institute{University of Galway, Ireland\\
\email{J.Lu5@nuigalway.ie}\\\email{patrick.mannion@universityofgalway.ie}\\\email{karl.mason@universityofgalway.ie}\\}
\maketitle              % typeset the header of the contribution
\begin{abstract}
Reinforcement learning is commonly applied in residential energy management, particularly for optimizing energy costs. However, RL agents often face challenges when dealing with deceptive and sparse rewards in the energy control domain, especially with stochastic rewards. In such situations, thorough exploration becomes crucial for learning an optimal policy. Unfortunately, the exploration mechanism can be misled by deceptive reward signals, making thorough exploration difficult. Go-Explore is a family of algorithms which combines planning methods and reinforcement learning methods to achieve efficient exploration. We use the Go-Explore algorithm to solve the cost-saving task in residential energy management problems and achieve an improvement of up to 19.84\% compared to the well-known reinforcement learning algorithms.

\end{abstract}
\section{Introduction}
% In the year 2021, residential structures were responsible for approximately 17\% of both direct and indirect carbon emissions, along with 21\% of total energy consumption on a global scale \cite{UN energy}. Minimizing the energy requirements and subsequent carbon footprint is of utmost importance in aligning with the carbon emissions reduction commitments made by governments globally. Renewable energy is widely recognized as a highly favorable solution for enhancing energy efficiency. Despite the projected significant expansion of global renewable energy capacity, reaching approximately 2400 gigawatts, the intermittent nature of renewable generation remains a challenge \cite{renewable}. An autonomous energy management technique that can  specifically work in the context of renewable energy generation is therefore a solution to handle this challenge.

Reinforcement learning (RL) has been widely used in autonomous energy control problems \cite{glavic2017reinforcement,huang2022mixed,ilager2020thermal,ren2022novel,shuvo2022home}. It is a paradigm where the agent learns from the interaction with the environment and solves the optimal decision-making problem. It only needs a reward function to give feedback from the environment as the evaluation of its behavior, therefore it is crucial that reward functions are well designed. 

In most RL applications in residential energy management, saving energy cost is always one of the most important tasks \cite{haq2022implementation,lu2022multi,xu2020multi,yu2019deep}. The reward function can be therefore constructed as feedback on the cost or the energy consumption. However, a raw cost reward function, e.g., directly using the cost as a reward, can lead the agent to local optima especially when intermittent renewable energy generation is incorporated. This is because such reward is highly stochastic and deceptive and the limit of the exploration mechanism cannot guarantee thorough exploration. The agent may occasionally find a relatively low price time interval and stay there forever even if there is a better price interval in the future as it may think this area in state space is the global optimum. 

We use the cutting-edge algorithm Go-Explore \cite{ecoffet2019go,ecoffet2021first}, which combines planning and RL to achieve efficient exploration for the agent to find and robustify the policy in an environment with stochastic raw reward signals.
To the best of our knowledge, this is the first application of the Go-Explore algorithm in energy management problems. our experimental results show that the Go-Explore agent surpasses the performance of the baseline RL algorithms.

\section{Background Knowledge}
\subsection{Markov Decision Process and Reinforcement Learning}
A sequential control task is always modeled as a Markov decision process (MDP). MDPs are defined with a tuple $(\mathcal{S},\mathcal{A},\mathcal{T},\mathcal{R},\gamma)$ \cite{sutton2018reinforcement}. $\mathcal{S}$ and $\mathcal{A}$ are the state space and action space. They are the set of all possible situations the RL can see in the environment and all actions it can take. $\mathcal{T}$ is the transition dynamics of the environment. $\mathcal{R}$ is the reward function that defines the task and the feedback to the RL agent. $\gamma$ is the discount factor for the agent to determine the importance of the long-term return.

RL is a method capable of solving the MDP when the transition $\mathcal{T}$ is unknown or partially known. In this work, we focus on model-free RL, where the agent learns through trial and error. This can be done either directly in a policy-based paradigm, indirectly in a value-based paradigm, or through a combination of both in an actor-critic paradigm. While different paradigms have been proposed, the ultimate goal remains the same: to maximize the cumulative reward. RL aims to learn the optimal policy that leads to the highest achievable cumulative reward by iteratively improving its behavior through interactions with the environment.

% \subsection{Reinforcement Learning for Residential Energy Management}

\subsection{Go-Explore}
While RL is supposed to be able to solve sequential control problems. It often struggles to learn an optimal policy when the environment is too complex to explore and the reward signal is sparse and deceptive. Ecoffet et al. mentioned that a thorough exploration can be the solution\cite{ecoffet2021first}. They pointed out that two main challenges in achieving efficient exploration are the phenomenon referred to as "detachment" and "derailment." These are the agent's tendency to forget how to return to previously discovered promising states, i.e. detachment, and failing to first return to the promising state and then start exploration from it, i.e. derailment. To handle those two challenges, they proposed a family of algorithms "Go-Explore" to simply memorize the promising cell (a cohesion of similar states) and firstly return to the state before exploration \cite{ecoffet2019go,ecoffet2021first}. This means that the agent can always remember "good" cells. To be able to return to those cells, the Go-Explore algorithm requires a simulator that is capable of being reset to a specific cell. This feature allows the agent to revisit and explore promising cells more efficiently. There are two phases of the Go-Explore algorithm.\\
\textbf{Phase 1: Explore until solve}
The phase starts by sampling the initial cell from the archive and starts exploration. The Go-Explore agent explores the environment as a usual episode iteration and stores promising cells in the archive along the trajectory it goes through. When an episode ends, a new cell is sampled from the archive and the simulator is reset to one state in the cell and the agent starts exploring from it. During the new round of exploration, if it finds any new promising cell, it will store it in the archive. If the agent finds any better trajectory to an existing cell, it will update the archive. This process is repeated until the problem is solved. Note that the solution is not guaranteed to be optimal and still needs further robustification to improve it.

\textbf{Phase 2: Robustification}
After successfully finishing phase 1, there should be some high-standard trajectories. However, these trajectories are not optimal due to the stochasticity of the environment and the related general policy is yet learned. Phase 2 is to learn a policy that is able to imitate the same routine of the trajectory of the agent in Phase 1 and improve upon it. 

\section{Model}
\subsection{Residential Energy Consumption Model}
To effectively manage residential energy loads, it is advantageous to categorize household appliances into different types. Residential loads can be classified into three categories mentioned in literature \cite{lu2022multi,lu2019demand}.
\begin{enumerate} 
    \item Shiftable loads.\\
    These are loads that can be rescheduled to take advantage of cheaper energy costs. Their operation time can be adjusted to align with periods of lower electricity prices.
    \item Non-shiftable loads.\\
    Non-shiftable loads encompass essential appliances that cannot be rescheduled or turned off, such as refrigerators and alarm systems. They operate continuously and require a consistent power supply.
    \item Controllable loads.\\
    Controllable loads refer to appliances where the power consumption can be flexibly adjusted by the user. This category includes appliances like air conditioners and lighting systems.
\end{enumerate}
In the scheduling process, a reinforcement learning (RL) agent is employed to manage the shiftable loads, while the non-shiftable and controllable loads are referred to as "background loads". It is assumed that shiftable appliances operate at their rated power when switched on.

Household energy demand draws power from both the grid and renewable generation. To minimize costs, the agent prioritizes the utilization of energy sourced from renewable sources when it is available. The grid adopts a dynamic price scheme, see price detail in Section \ref{subsec:Datasets}. 

In this study, the selected shiftable appliance is the "Bosch WAJ28008GB Washing Machine," rated at 1 kW. We assume that the washing machine needs to operate for 2 hours per day. If the agent fails to run the washing machine throughout the day, it will be compelled to operate during the final 2 hours of the day.

\subsection{Markov Decision Process Setup}
In this section, we will present the construction of the MDP of this work.
\subsubsection{State Space}
The state space comprises several variables, including:
\begin{enumerate}
    \item Price: Represents the last hour's dynamic average price of electricity from the grid.
    \item Renewable generation: The amount of the average renewable generation in the last hour.
    \item Background loads: Refers to the average background loads of the last hour.
    \item Remaining task: The number of hours left for the shiftable load to operate.
    \item Time by hour: Represents the current hour of the day.
\end{enumerate}
\subsubsection{Action Space}
The action is a binary choice, where 0 is for not running and 1 is for running. If action 1 is picked, the appliance will start working based on the price, background loads, and renewable generation of this hour.
\subsubsection{Reward Function}
This is a single-objective RL problem, therefore the only reward function we used is the hourly cost. 
\begin{equation}
    r_{t} = -price_{t}\cdot max[p^{s}_{t}+p^{b}_{t}-p^{r}_{t},0]
\end{equation}
where $price_{t}$ is the electricity price at time $t$, $p^{s}_{t}$ is the power of shiftable loads, $p^{b}_{t}$ and $p^{r}_{t}$ is the power of background loads and renewable generation separately. The reward is calculated by multiplying the price of electricity at time t by the maximum value between ($p^{s}_{t} + p_t^b - p_t^r$) and 0. This formulation encourages minimizing the total power consumption and maximizing the utilization of renewable generation, as a higher cost for the maximum term will result in a higher penalty.
\subsection{Go-Explore Model}
The details of the Go-Explore model are presented in this section. 
The Go-Explore archive stores three types of entries, e.g. the cell, the trajectory leads to the cell, and cell-related information, i.e. a tuple $(number\ of\ visits, cost)$. $number\ of\ visits$ is the number of how many times this cell is visited, and $cost$ is the energy cost of the trajectory. Each cell representation is the tuple $(remaining\ task, time\ by\ hour)$.
\subsubsection{Phase 1: Explore until solve}
\begin{enumerate}
    \item \textbf{Sample a cell from the archive:} There is only one initial cell stored in the archive. As the episode proceeds, there are more cells added to the archive. Then the cell will be sampled with the probability calculated from the scores. In this work, we use a plain score that is the reciprocal of the number of visits. All cells' score is normalized to fit in the [0,1] interval as a probability distribution.
    \item \textbf{Explore from the sampled cell:} The simulator is reset to the sampled cell, and randomly samples actions as exploration from that cell.
    \item \textbf{Update the archive:} The archive is updated if a better trajectory is found or a new cell is found. In this work, a better trajectory is the trajectory that resulted in less cost to reach the cell than the original trajectory.
    \item \textbf{Repeat} the aforementioned three steps until the problem is solved.
\end{enumerate}
We separate the robustification phase into two parts: policy cloning and robustification. This is to provide a training process on a higher granularity.\\
\textbf{Phase 2.1: Policy Cloning}
A PPO agent is trained by simply imitating the demonstration from Phase 1. It does not know anything about the true reward signal but will receive a reward of 1 if the next state is aligned with the demonstration, otherwise, the reward is 0. The agent trained in this phase is noted as "Go-Explore (no robustification)".\\
\textbf{Phase 2.2: Robustification}
With the policy cloning agent, we further train it with a true reward signal. The agent trained in this phase is noted as "Go-Explore (robustification)".

\section{Experiment}
\subsection{Datasets}
\label{subsec:Datasets}
We use two datasets used in the work of Lu et al. \cite{lu2022multi}. 
\begin{itemize}
    \item \textbf{Electricity Price:} 
    The dataset for electricity prices is sourced from the PJM dataset \cite{2021PJMDataset}. The training data covers the period from 01/05/2021 00:00 to 02/05/2021 00:00. The evaluation data spans one month, starting from 01/05/2021 00:00 to 31/05/2021 00:00.
    
    \item \textbf{Background Load and Renewable Generation: } 
    The renewable generation and background load is sourced from the Smart* dataset for Sustainability within the "Home C" of UMass dataset \cite{barker2012smart}. The training data covers the period from 01/05/2014 00:00 to 02/05/2014 00:00. The evaluation data spans one month, starting from 01/05/2014 00:00 to 31/05/2014 00:00.
\end{itemize}

\subsection{Baseline Algorithm}
We two renowned RL algorithms as the baseline, i.e. proximal policy optimization (PPO) \cite{schulman2017proximal} and deep Q-network (DQN)\cite{mnih2015human}. Both algorithms have been instrumental in solving challenging decision-making problems in diverse domains. We also use the PPO algorithm for the policy cloning and robustification phase in our Go-Explore implementation that shares the same hyperparameters as the pure PPO agent. 
The learning rates for PPO and DQN are all 0.001, while the discount factor is 1 and the batch size is 64. The other hyperparameters are detailed in Table \ref{tab:DQN Training Hyperparameters}.
\begin{table}[h]
\centering
\caption{Hyperparameters}
\label{tab:DQN Training Hyperparameters}
{\begin{tabular}{c|c|c|c|c}
    \hline 
    Alg.&Number of Episodes&Hidden Layer&KL-target&Entropy Weight\\
    \hline   
    PPO&60&[32,32,32]&0.01&0.001\\
    DQN&5000&[32,32,16]&-&-\\
    \hline
\end{tabular}}
\end{table}
\section{Result and Discussion}
Table \ref{tab: result} presents the results of the Go-Explore algorithm and the baselines. 
The monthly cost values are denoted in Euro (\EUR). The cost-saving column specifically compares the results of the "Go-Explore (robustification)" simulation with the other three simulations. It showcases the absolute value of cost saving and the relative improvement of cost saving compared to the DQN-agent cost.
\begin{table}
\centering
\caption{Experiment Result}\label{tab: result}
\begin{tabular}{c|c}
\hline
Algorithm &  Cost Saving vs. DQN-agent (\EUR 95.65) \\
\hline
% DQN &  - & 19.84\%\\
PPO &  \EUR 16.49 (17.23\%)\\
Go-Explore (no robustification) & \EUR18.97 (19.83\%)\\
Go-Explore (robustification) & \EUR18.98 (19.84\%)\\
\hline
\end{tabular}
\end{table}
The Go-Explore (robustification) agent saves \EUR18.98 which reduced 19.84\% cost than the DQN agent, achieving the highest saving in simulations. It is noted that the improvement between the versions with/without robustification is very close. This is because the environment is relatively deterministic as the action can only influence a limited number of states so the agent cannot exploit many benefits in the robustification phase. However, if the stochasticity of the transition increase, e.g. in a multi-agent setting, a robustification version can be better.

\section{Conclusion}
We use the Go-Explore algorithm to solve the cost-saving task in residential energy management and have achieved a cost-saving of up to 19.84\%. The combination of planning and RL is promising in hard-exploration real-life problems. Future extensions of this work can be:\\
- Apply the "policy-based Go-Explore" \cite{ecoffet2021first} to improve the training efficiency .\\
- Extension of the environment to a multi-agent environment.

% \subsubsection{Acknowledgements} This research is funded by the Government of Ireland Postgraduate Scholarship (GOIPG/2022/2140).

\bibliographystyle{splncs04}
\bibliography{sn-bibliography}

\begin{thebibliography}{10}
\providecommand{\url}[1]{\texttt{#1}}
\providecommand{\urlprefix}{URL }
\providecommand{\doi}[1]{https://doi.org/#1}

\bibitem{barker2012smart}
Barker, S., Mishra, A., Irwin, D., Cecchet, E., Shenoy, P., Albrecht, J.,
  et~al.: Smart*: An open data set and tools for enabling research in
  sustainable homes. SustKDD, August  \textbf{111}(112), ~108 (2012)

\bibitem{ecoffet2019go}
Ecoffet, A., Huizinga, J., Lehman, J., Stanley, K.O., Clune, J.: Go-explore: a
  new approach for hard-exploration problems. arXiv preprint arXiv:1901.10995
  (2019)

\bibitem{ecoffet2021first}
Ecoffet, A., Huizinga, J., Lehman, J., Stanley, K.O., Clune, J.: First return,
  then explore. Nature  \textbf{590}(7847),  580--586 (2021)

\bibitem{glavic2017reinforcement}
Glavic, M., Fonteneau, R., Ernst, D.: Reinforcement learning for electric power
  system decision and control: Past considerations and perspectives.
  IFAC-PapersOnLine  \textbf{50}(1),  6918--6927 (2017)

\bibitem{haq2022implementation}
Haq, E.U., Lyu, C., Xie, P., Yan, S., Ahmad, F., Jia, Y.: Implementation of
  home energy management system based on reinforcement learning. Energy Reports
   \textbf{8},  560--566 (2022)

\bibitem{huang2022mixed}
Huang, C., Zhang, H., Wang, L., Luo, X., Song, Y.: Mixed deep reinforcement
  learning considering discrete-continuous hybrid action space for smart home
  energy management. Journal of Modern Power Systems and Clean Energy
  \textbf{10}(3),  743--754 (2022)

\bibitem{ilager2020thermal}
Ilager, S., Ramamohanarao, K., Buyya, R.: Thermal prediction for efficient
  energy management of clouds using machine learning. IEEE Transactions on
  Parallel and Distributed Systems  \textbf{32}(5),  1044--1056 (2020)

\bibitem{lu2022multi}
Lu, J., Mannion, P., Mason, K.: A multi-objective multi-agent deep
  reinforcement learning approach to residential appliance scheduling. IET
  Smart Grid  \textbf{5}(4),  260--280 (2022)

\bibitem{lu2019demand}
Lu, R., Hong, S.H., Yu, M.: Demand response for home energy management using
  reinforcement learning and artificial neural network. IEEE Transactions on
  Smart Grid  \textbf{10}(6),  6629--6639 (2019)

\bibitem{mnih2015human}
Mnih, V., Kavukcuoglu, K., Silver, D., Rusu, A.A., Veness, J., Bellemare, M.G.,
  Graves, A., Riedmiller, M., Fidjeland, A.K., Ostrovski, G., et~al.:
  Human-level control through deep reinforcement learning. nature
  \textbf{518}(7540),  529--533 (2015)

\bibitem{2021PJMDataset}
PJM: 2021 pjm dataset.
  \url{https://dataminer2.pjm.com/feed/rt\_fivemin\_mnt\_lmps} (2021),
  \url{https://www.pjm.com/markets-and-operations}

\bibitem{ren2022novel}
Ren, M., Liu, X., Yang, Z., Zhang, J., Guo, Y., Jia, Y.: A novel forecasting
  based scheduling method for household energy management system based on deep
  reinforcement learning. Sustainable Cities and Society  \textbf{76},  103207
  (2022)

\bibitem{schulman2017proximal}
Schulman, J., Wolski, F., Dhariwal, P., Radford, A., Klimov, O.: Proximal
  policy optimization algorithms. arXiv preprint arXiv:1707.06347  (2017)

\bibitem{shuvo2022home}
Shuvo, S.S., Yilmaz, Y.: Home energy recommendation system (hers): A deep
  reinforcement learning method based on residents’ feedback and activity.
  IEEE Transactions on Smart Grid  \textbf{13}(4),  2812--2821 (2022)

\bibitem{sutton2018reinforcement}
Sutton, R.S., Barto, A.G.: Reinforcement learning: An introduction. MIT press
  (2018)

\bibitem{xu2020multi}
Xu, X., Jia, Y., Xu, Y., Xu, Z., Chai, S., Lai, C.S.: A multi-agent
  reinforcement learning-based data-driven method for home energy management.
  IEEE Transactions on Smart Grid  \textbf{11}(4),  3201--3211 (2020)

\bibitem{yu2019deep}
Yu, L., Xie, W., Xie, D., Zou, Y., Zhang, D., Sun, Z., Zhang, L., Zhang, Y.,
  Jiang, T.: Deep reinforcement learning for smart home energy management. IEEE
  Internet of Things Journal  \textbf{7}(4),  2751--2762 (2019)

\end{thebibliography}
% \bibitem{UN energy}
% United Nations Environment Programme (2022). 2022 Global Status Report for Buildings and Construction: Towards
% a Zero‑emission, Efficient and Resilient Buildings and Construction Sector. Nairobi.  https://globalabc.org/

% \bibitem{renewable}
% IEA (2022), Renewables 2022, IEA, Paris https://www.iea.org/reports/renewables-2022, License: CC BY 4.0

% \bibitem{Reinforcement learning for electric power system decision and control}
% Glavic, M., Fonteneau, R., \& Ernst, D. (2017). Reinforcement learning for electric power system decision and control: Past considerations and perspectives. IFAC-PapersOnLine, 50(1), 6918-6927.

% \bibitem{Energy management for a hybrid electric vehicle based on prioritized deep reinforcement learning framework}
% Du, G., Zou, Y., Zhang, X., Guo, L., \& Guo, N. (2022). Energy management for a hybrid electric vehicle based on prioritized deep reinforcement learning framework. Energy, 241, 122523.
% \end{thebibliography}
\end{document}